\documentclass[sigconf,anonymous=false,nonacm,authorversion]{acmart}

\usepackage{soul}
\newcommand{\drop}[1]{\textcolor{red}{#1}}
\renewcommand{\drop}[1]{}
\definecolor{cadmiumgreen}{rgb}{0.0, 0.42, 0.24}
\usepackage{multirow}
\usepackage{xcolor}
\usepackage{enumitem}

\usepackage{graphicx}
\usepackage{booktabs} 
\usepackage{pifont}
\usepackage{mathtools}

\usepackage[ruled, vlined, norelsize]{algorithm2e}
\SetKwInput{KwInput}{Input}
\SetKwInput{KwOutput}{Output}
\SetKwInput{KwInit}{Initialization}
\SetKwInput{Kwprocedure}{Procedure}
\usepackage{lipsum}
\usepackage{dblfloatfix}
\usepackage{bm}

\definecolor{cadmiumgreen}{rgb}{0.0, 0.42, 0.24}
\newcommand{\cmark}{\ding{51}}
\newcommand{\xmark}{\ding{55}}

\usepackage{listings}
\lstdefinelanguage{Verilog}{
  morekeywords={module, endmodule, input, output, reg, wire, always, begin, end, if, else, for, while, case, default},
  sensitive=false,
  morecomment=[l]{//},
  morecomment=[s]{/*}{*/},
  morestring=[b]",
}
\definecolor{shadecolor}{rgb}{0.9,0.9,0.9}
\usepackage{acronym}
\acrodef{IC}{integrated circuit}
\acrodef{EDA}{electronic design automation}
\acrodef{HDL}{hardware description language}
\acrodef{ML}{machine learning}
\acrodef{IP}{intellectual property}
\acrodef{RTL}{register transfer level}
\acrodef{QoR}{quality-of-result}
\acrodef{DNN}{deep neural network}
\acrodef{MDP}{Markov decision process}
\acrodef{CNN}{convolutional neural network}
\acrodef{RL}{reinforcement learning}
\acrodef{MSE}{mean-square error}
\acrodef{SA}{simulated annealing}
\acrodef{SOTA}{state-of-the-art}
\acrodef{LLM}{large language model}
\acrodef{PPA}{power performance area}
\acrodef{MCTS}{Monte-Carlo tree search}

\usepackage{amsmath,amsfonts,bm}

\def\eqref#1{equation~\ref{#1}}

\def\1{\bm{1}}

\DeclareMathAlphabet{\mathsfit}{\encodingdefault}{\sfdefault}{m}{sl}
\SetMathAlphabet{\mathsfit}{bold}{\encodingdefault}{\sfdefault}{bx}{n}

\DeclareMathOperator*{\argmax}{arg\,max}

\begin{document}

\title{Make Every Move Count: LLM-based High-Quality RTL Code Generation Using MCTS}

\begin{abstract}
Existing large language models (LLMs) for register transfer level code generation face challenges like compilation failures and suboptimal power, performance, and area (PPA) efficiency. This is due to the lack of PPA awareness in conventional transformer decoding algorithms. In response, we present an automated transformer decoding algorithm that integrates Monte Carlo tree-search for lookahead, guiding the transformer to produce compilable, functionally correct, and PPA-optimized code. Empirical evaluation with a fine-tuned language model on RTL codesets shows that our proposed technique consistently generates functionally correct code compared to prompting-only methods and effectively addresses the PPA-unawareness drawback of naive large language models. 
For the largest design generated by the state-of-the-art LLM (16-bit adder), our technique can achieve a 31.8\% improvement in the area-delay product.
\end{abstract}


\author{Matthew DeLorenzo}
\email{matthewdelorenzo@tamu.edu}
\affiliation{
    \institution{Texas A\&M University}
    \city{College Station}
    \state{Texas}
    \country{United States}
}

\author{Animesh Basak Chowdhury}
\email{abc586@nyu.edu}
\affiliation{
    \institution{New York University}
    \city{New York}
    \state{New York}
    \country{United States}
}

\author{Vasudev Gohil}
\email{gohil.vasudev@tamu.edu}
\affiliation{
    \institution{Texas A\&M University}
    \city{College Station}
    \state{Texas}
    \country{United States}
}

\author{Shailja Thakur}
\email{shailjathakur@nyu.edu}
\affiliation{
    \institution{New York University}
    \city{New York}
    \state{New York}
    \country{United States}
}

\author{Ramesh Karri}
\email{rkarri@nyu.edu}
\affiliation{
    \institution{New York University}
    \city{New York}
    \state{New York}
    \country{United States}
}

\author{Siddharth Garg}
\email{sg175@nyu.edu}
\affiliation{
    \institution{New York University}
    \city{New York}
    \state{New York}
    \country{United States}
}

\author{Jeyavijayan Rajendran}
\email{jv.rajendran@tamu.edu}
\affiliation{
    \institution{Texas A\&M University}
    \city{College Station}
    \state{Texas}
    \country{United States}
}
\maketitle

\section{Introduction}
\label{sec:Introduction}

Large language models (LLMs) have been a significant breakthrough in artificial intelligence, demonstrating remarkable success in solving various real-world problems. These models, trained on vast amounts of text data, have shown an uncanny ability to generate human-like text, understand context, answer questions, and even write code. They have been successfully deployed in numerous applications, including customer service, content creation, and language translation, to name a few~\cite{LLMs_applications_blog}. 
The versatility and robustness of LLMs have made them an invaluable tool in the AI toolkit, opening up avenues for exploration and innovation. Microsoft, Google, Meta, and Amazon have invested heavily in generative AI technologies such as LLMs~\cite{Google_Microsoft_Amazon_Meta_LLM_article}.

One such avenue that has garnered attention is the use of LLMs in chip design~\cite{thakur_benchmarking_2023,blocklove_chip-chat_2023,liu2023verilogeval,liu2023chipnemo}. Digital chip design, a complex and intricate process, involves the creation of integrated circuits 
used in various electronic devices. A crucial part of this process is the generation of Verilog RTL codes, which describe the behavior of these digital circuits. Recent advancements have seen the successful application of LLMs in automating this task, thereby potentially revolutionizing the chip design process~\cite{thakur_benchmarking_2023,blocklove_chip-chat_2023,thakur2023autochip,kande2023llm}.

However, the current use of LLMs in Verilog generation has limitations. 
Although these models produce functional RTL codes for some simple modules and outperform  ChatGPT (GPT-3.5-turbo) and GPT4, they surprisingly fail to generate correct codes for commonly used circuits such as 8-bit adders and multipliers.
Additionally, they overlook the optimization aspect of the generated codes, including considerations such as the number of gates, area, and delay, among other factors that are crucial in efficient chip design. Generating optimized RTL codes is a complex task that requires a balance between functionality and efficiency, a balance to which current LLMs are agnostic. 

We overcome this hurdle by designing an automated technique to explore the trade-offs between different competing objectives of optimized RTL generation using LLMs. 
In particular, we design a Monte Carlo tree-search (MCTS) algorithm over the tokens produced by the LLM. 
Our approach aims to generate optimized RTL codes that function as intended and adhere to the principles of efficient chip design as the user desires. 
By exploring the vast search space 
(exponential in the number of tokens generated by the LLM) 
of potential token combinations,
our MCTS-based approach aims to find an optimal RTL based on the user's requirements while still ensuring functional correctness.

However, integrating this MCTS-based approach with LLMs presents unique challenges.
(i) One of the main issues we encounter is related to search efficiency. The vast number of potential token combinations produced by the LLM makes the search space extremely large, posing a significant challenge to the efficiency of the MCTS. (ii) Moreover, this issue is further exacerbated 
by the need to evaluate each potential combination, i.e., RTL code, for compilability, functionality, and performance metrics such as area, delay, etc. --- a task requiring significant computational resources. 
(iii) Finally, unlike prior works, we target generating practical designs such as 64-bit adders, which require an extremely large number of tokens and thus time.
We overcome these challenges by modifying our MCTS approach to (i) reduce the search complexity, (ii) incorporate feedback from synthesis tools, and (iii) leverage modularity to reuse our MCTS-generated optimized codes to work with practical designs.
Our final optimized MCTS approach allows us to use an off-the-shelf LLM to generate Verilog codes with desired characteristics (e.g., less delay or area) with much higher accuracy than prior works, which simply prompt the LLM to generate the code.
The main contributions of this work are:
\begin{enumerate}
    \item To the best of our knowledge, we devise the first technique to enhance LLMs for Verilog generation using MCTS.
    \item We overcome challenges with search space and scalability to practical designs using domain-specific optimizations.
    \item Unlike prior works, our approach generates functionally correct Verilog codes for various designs such as adders, multipliers, and multiply-accumulate (MAC) units.
    \item We are the first to leverage MCTS formulation to produce PPA-optimized codes to meet user objectives using LLMs.
\end{enumerate}
\section{Background and related work}
\label{sec:background}

\subsection{Large Language Models for Code Generation}

In recent advancements in code generation, modern transformer-based language models like BERT~\cite{devlin-etal-2019-bert}, GPT-2~\cite{radford2019language}, and T5~\cite{2020t5} have revolutionized the treatment of programming languages, drawing inspiration from their success in natural language tasks. A notable family of BERT-based transformers has emerged, specializing in code syntax task~\cite{kanade2020learning,feng2020codebert,devlin-etal-2019-bert,guo2020graphcodebert}. Building upon this, CodeX~\cite{chen_evaluating_2021} and CodeT5~\cite{wang_codet5_2021} have taken a step further, adopting GPT-2 and T5 as backbones for both code understanding and generation. A recent standout, AlphaCode~\cite{li2022competition}, has combined large transformer models pre-trained on extensive program data with large-scale sampling, demonstrating competitive performance in programming competitions. Notably, these efforts primarily concentrated on enhancing the capabilities of code-generation models, utilizing techniques like beam search~\cite{graves2012sequence}, where, instead of greedily choosing the most likely next token as the sequence is constructed, all possible next tokens are expanded, and the $k$ most likely are kept.\footnote{$k$ is a user-specified parameter and controls the number of beams or parallel searches through the sequence of probabilities.}

\subsection{LLMs for Verilog RTL Code Generation}
Recent works~\cite{thakur_benchmarking_2023,blocklove_chip-chat_2023,liu2023chipnemo} have used LLMs to generate RTL codes. DAVE~\cite{pearce_dave_2020} is one of the first attempts to use GPT-2 to translate English into RTL code. Recently, VeriGen~\cite{thakur_benchmarking_2023} demonstrated that CodeGen, an open-source LLM fine-tuned on RTL data from GitHub and textbooks, performed better than the code-davinci-002~\cite{chen_evaluating_2021} on 17 RTL tasks. VerilogEval~\cite{liu2023verilogeval} introduced a benchmark of $156$ RTL coding challenges and showed that the performance of pre-trained LLMs on RTL code generation could be enhanced by supervised fine-tuning with synthetic problem-code pairs created by LLM. ChipNeMo~\cite{liu2023chipnemo} demonstrated that open-source LLMs like LLaMa2 7B/13B can be fine-tuned for domain-specific objectives like RTL code generation, script generation for EDA tools, and bug summarization. RTLLM~\cite{lu2023rtllm} presented prompt engineering methods to improve the quality of RTL generation on a set of open-source hardware designs. Chip-Chat~\cite{blocklove_chip-chat_2023} used conversational interfaces to design and verify an 8-bit accumulator-based microprocessor with GPT-4 and GPT-3.5. It reported that GPT-4 generated codes of relatively high quality, but it was still not good enough at detecting and correcting errors. Recently, AutoChip~\cite{thakur2023autochip} improved on~\cite{thakur_benchmarking_2023} by using compilation errors from LLM-generated RTL codes as feedback to improve code generation performance. To the best of our knowledge, no prior work explicitly focuses on LLM decoding algorithms to improve RTL code generation performance in terms of (i) functional correctness and (ii) PPA optimization. In fact, as evidenced by our results in Sec.~\ref{sec:results}, existing LLM-based approaches do not even yield a functional result for commonly used arithmetic modules (such as adders and multipliers taught in undergraduate classes) most of the time, let alone PPA-optimized results.
\section{Framework}\label{sec:framework}

Here, we formulate the problem of generating optimal Verilog codes using LLMs as a Markov decision process (MDP) and then devise a tree-search-based algorithm to solve the MDP. However, the preliminary formulation suffers from challenges related to efficiency and efficacy, which we analyze and address in Secs.~\ref{sec:challenge_sol_1} and~\ref{sec:challenge_sol_2}, resulting in our final formulation in Sec.~\ref{sec:putting_it_all_together}.

\subsection{Preliminary Formulation}
We consider the Verilog code generation problem, where the LLM is given a description of the Verilog coding problem along with the module definition containing the input and output ports. Given this input prompt, the LLM is expected to generate a Verilog code that satisfies the functionality described in the input prompt. Prior works on LLMs for Verilog generation have only shown reasonable performance for generating functionally correct outputs for simple modules~\cite{thakur_benchmarking_2023}. However, their performance drops when generating slightly more advanced and practical codes such as adders and multipliers.
Additionally, the chip design process requires generating Verilog code with not only the correct functionality but also with desired trade-offs between competing objectives such as area and delay. As we show in Sec.~\ref{sec:results}, existing LLMs fail to achieve the desired trade-offs. To alleviate this issue, we augment the target LLM with a decision-making agent to optimize the user-defined objective such as area, delay, or area-delay product. To this end, we formulate an MDP
that, when solved by the decision-making agent along with tokens generated by the LLM, yields a functionally correct Verilog code that optimizes the user-defined objective.

\begin{itemize}[leftmargin=*]
\item \textbf{State $S_t$} at step $t$ is the prompt of problem description combined with partial RTL code $P_{t}$ represented as a token sequence. $S_0$ represents the initial state, representing only the prompt. We define terminal state $S_T$ as either representing an RTL code having terminal token \texttt{endmodule} or when $t=T_{max}$ tokens have been generated.

\item \textbf{Actions $\mathcal{A}$} is the vocabulary set of tokens of LLM. An individual action $a_t$ is the next token chosen by our policy for exploration.

\item \textbf{State transition $\tau(S_{t+1}|S_t,a_t)$} is the probability that action $a_t$ in state $S_t$ leads to the state $S_{t+1}$. 
In our case, the transition function is deterministic. Chosen action $a_t$ is appended to state $s_t$ to generate  next state: $S_{t+1} = S_t \circ a_t$, 
where $\circ$ is concatenation.

\begin{figure*}[t]
    \centering
    \includegraphics[width=\textwidth]{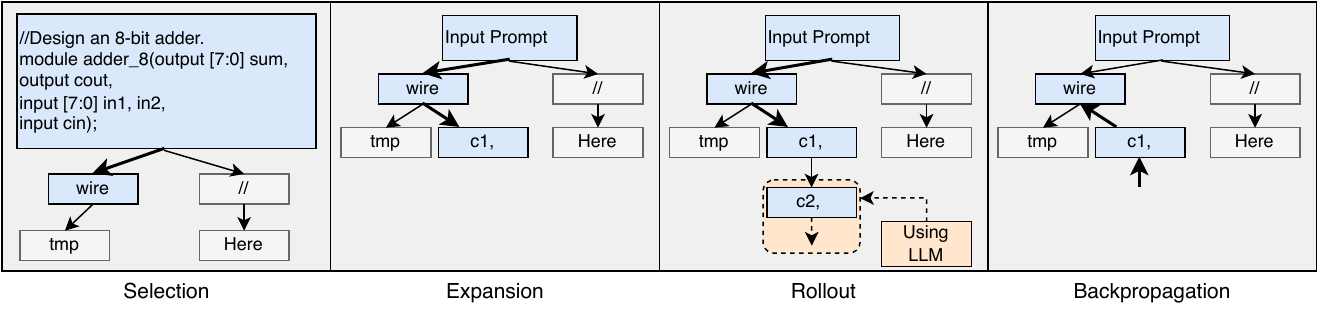}
    \caption{Illustrative example for the different phases in each MCTS iteration with an 8-bit adder module.}
    \label{fig:MCTS_adder_example}
\end{figure*}

\item \textbf{Reward function $\mathcal{R}$ ($S_{t}$)} is determined based on the syntactical correctness, functional correctness, and \ac{PPA} metric of the design. We consider the inverse area-delay product (larger the better) for the PPA score, and  $\mathcal{R}$ becomes
\begin{equation}
    \mathcal{R}(S_t)=
    \begin{dcases}\label{eq:1}
        0, & \text{if } S_{t}\neq S_T \text{ or } t\neq T_{max}\\
        \alpha_{NC}, & \text{else if } S_{t} \text{ is not compilable}\\
        \alpha_{NF}, & \text{else if } S_{t} \text{ is not functional}\\
        \alpha_B + \left(1-\frac{a\times d}{a^{\dagger} \times d^{\dagger}}\right), & \text{otherwise}
    \end{dcases}
\end{equation}
where $\alpha_{NC}$ (<$0$) and $\alpha_{NF}$ (<$0$) are penalties for codes that are not compilable and not functional, respectively; $\alpha_B$ (>$0$) is a baseline reward for a functional code; $a$ and $d$ denote the area and delay of the synthesized netlist obtained from $S_{t}$, respectively; and $a^{\dagger}$ and $d^{\dagger}$ denote the area and delay of the first synthesized netlist for the current module. The reward is designed so that the solution of the MDP yields a Verilog code in line with the user's objective, which is a minimal area-delay product (ADP) in this case.
\end{itemize}

Thus, we seek to solve the following optimization problem:
\begin{equation}
\argmax_{P_{T} \in \mathcal{A}^{T_{max}}}  \mathcal{R}(S_T), \, \, s.t. \, \, S_{t+1} = S_{t} \circ a_{t} \, \, \forall t\in [0,T_{max}-1].
\end{equation}

To this end, we use an MCTS algorithm to find the optimal solution for this MDP, as explained next.

\subsection{LLM-based Code Generation Using MCTS}\label{sec:mcts_explanation}
The MCTS agent starts with the initial state $S_0$ (root node), which is the sequence of tokens representing the initial prompt describing the high-level specification of the module to design. From a given state $S_t$, the agent uses a policy, $\pi$, to pick the next token, i.e., an action $a_t \in \mathcal{A}$, and arrives at the next state, $S_{t+1}$, which is the total sequence of tokens so far (with new token appended). The agent receives a delayed reward $\mathcal{R}(S_T)$ on reaching terminal state $S_T$. At the conclusion of each MCTS iteration, the algorithm updates two parameters for each state $S_t$ along its path: (i) $N(S_t)$: visit count of state $S_t$ and (ii) $M(S_t)$: the total sum of rewards obtained by exploring all terminal states from state $S_t$.

The policy $\pi(S_t)$, guiding the MCTS token decision at each state, 
compares each potential next action.
For each action, $a_t$, this is done by utilizing the total sum of rewards value for the corresponding state, $N(S_t,a_t)$, and the visit count of the corresponding state, $M(S_t,a_t)$. The average reward is found through these parameters, resulting in the exploitation term to encourage high reward paths. The policy balances this exploitation term with the Upper Confidence Tree (UCT) term, encouraging the visitation of less visited states (exploration), which includes the LLM's probability value of choosing the action ($P(S_t,a_t)$). These two terms are used as follows:
\begin{equation}
\pi(S_t) = \argmax_{a_t \in \mathcal{A}} \left(\underbrace{\frac{M(S_t,a_t)}{N(S_t,a_t)}}_{\text{Avg. reward}} + c_{PUCT}\times{P(S_t,a_t)}{\underbrace{\frac{\sqrt{1 + N(S_t)}}{1 + N(S_t,a_t)}}_{\text{UCT Term}}}\right)
\label{eq:search-policy}
\end{equation}
where $c_{PUCT}$ denotes a constant exploration factor~\cite{kocsis2006bandit}.

We now detail the MCTS algorithm for an example 8-bit adder circuit using Figure~\ref{fig:MCTS_adder_example}. Each iteration consists of four stages: (i) selection, (ii) expansion, (iii) rollout, and (iv) backpropagation.
During selection, a search tree is built from the initial state (the input prompt in Figure~\ref{fig:MCTS_adder_example}) by following a search policy (Eq. (\ref{eq:search-policy})), with the aim of balancing exploration and exploitation.
The selection phase repeats until the chosen action (child node) has not been visited yet and, therefore, has no corresponding node in the current MCTS tree. At this point, the expansion phase occurs in which a new node is created for that chosen action. Then the rollout phase occurs, in which the next token is determined by the max probability estimates by the LLM. This rollout phase continues (LLM choosing the most likely next token) until a terminal state has been reached.
In the next phase, backpropagation, the reward, $\mathcal{R}(S_T)$, is first determined for the final state (the generated Verilog code) according to Eq. (\ref{eq:1}). This obtained reward is then backpropagated in the tree such that the values of each node (total reward $M(\cdot)$ and visit counts $N(\cdot)$) in the explored path (starting from the expanded node) are updated accordingly until the root node is reached. The next MCTS iteration can then occur, in which the selection phase utilizes these updated values in its policy.

Our experiments indicate that while the formulation described above generates functional Verilog codes for smaller modules, it faces two challenges, particularly when working with codes that require a large number of tokens, i.e., when number of time steps is large. 
Next, we analyze the challenges and devise solutions.

\subsection{Reducing Search Space}\label{sec:challenge_sol_1}

\begin{figure*}[t]
    \centering
    \includegraphics[width=\textwidth]{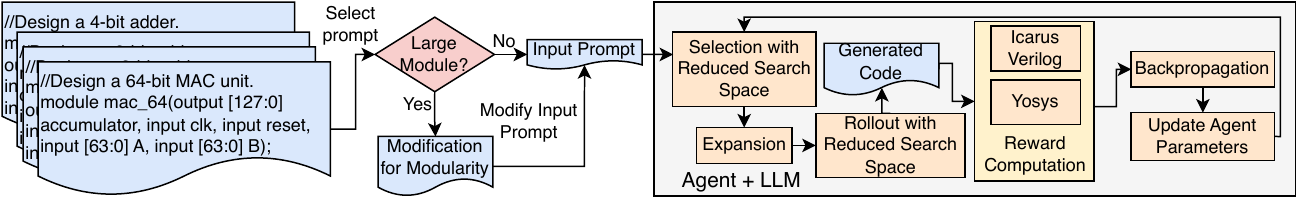}
    \caption{Final MCTS framework}
    \label{fig:final_framework_figure}
\end{figure*}

\noindent\textbf{Challenge 1: Large Search Space.} In the above formulation, each new action (i.e., token) results in a new state for the agent. This means that the tree depth increases with each time step, leading to a large search space to optimize over. For instance, if, at each time step, we only consider the top $k$ next tokens as our possible actions, the number of paths for the MCTS algorithm is $k^{T}$. Here, $T$ is the number of time steps, i.e., the number of tokens for that generated code, which can be in the order of several hundreds, if not thousands, leading to an immense search space for the MCTS algorithm. Hence, for a fixed number of iterations, a small portion of the tree is explored, or equivalently, exploring a large portion of the tree requires many iterations and a larger runtime. 

\noindent\textbf{Solution 1.} To tackle the large search space, we introduce a method that prunes unnecessary paths from consideration for the MCTS. We analyze each new action (i.e., token chosen by the agent) and check whether it is part of the functional code for the given module. This is done because LLMs generate tokens that do not affect the functionality of the Verilog code but are added to enhance clarity for a human reader. Examples of non-functional tokens are those that are part of comment lines/phrases. Tokens present in such comment lines/phrases increase the number of paths for MCTS without impacting the functional correctness of the code or the performance (e.g., delay/area). Hence, we remove them from consideration for our MCTS algorithm. By doing so, we reduce the number of tokens, $T$, thereby reducing the number of paths (because of its exponential influence on the number of paths).

In implementing this feature of removing unnecessary comment paths in MCTS, two stages are altered: selection and rollout. 
During selection, any comment-line tokens (``\texttt{//}'', ``\texttt{/*}'', or ``\texttt{*/}'') are removed from consideration so that they are not selected. Similarly, to prevent comments in the rollout stage, if the LLM selects a comment-line token as most likely, the next most likely token is evaluated and selected if it is not a comment-line token (continuing until a non-comment token is found).

\begin{table}[t]
\caption{Comparison of MCTS iteration rates per minute for adder modules for 4-bit, 64-bit without modularity (w/o Mod.), and 64-bit with modularity (w/ Mod.).}
\label{tab:challenge_sol2}
\resizebox{0.48\textwidth}{!}{
\begin{tabular}{cccc}
\toprule
& 4-bit & 64-bit (w/o Mod.) & 64-bit (w Mod.)\\ \midrule
\begin{tabular}[c]{@{}c@{}}MCTS Iteration\\  Rate / min\end{tabular} & 0.72 & 0.08 & 0.24 \\ \bottomrule
\end{tabular}
}\vspace{-0.2in}
\end{table}

\subsection{Improving Efficiency Through Modularity}\label{sec:challenge_sol_2}
\noindent\textbf{Challenge 2: Lack of Scalability to Large Modules.} Another crucial limitation of the preliminary formulation described above is its poor performance and inability to produce a functional or even a complete Verilog code for modules that require a large number of tokens, such as a 64-bit adder. 
This is because the time required to generate the next token increases as the number of generated tokens increases, essentially resulting in a lower MCTS iteration rate. In other words, the time required to generate codes for large modules increases with code size. 
Due to this, the rate of MCTS path exploration is significantly reduced, leading to poor performance.

\noindent\textbf{Solution 2.} To address this challenge, we restrict the number of tokens that need to be generated for larger modules by reusing the optimized modules found by our MCTS framework for the smaller modules. 
For instance, as generating a 64-bit adder requires a large depth of tokens to complete, resulting in a large runtime, we provide the context of previously generated sub-modules (such as an MCTS-generated 8-bit adder) via the design prompt, thereby enabling the LLM to reference them in its large-module generation. This allows the LLM generation to target the higher-level implementation and provides a more feasible MCTS iteration time, enabling deeper testing. 
Table~\ref{tab:challenge_sol2} demonstrates the impact of this solution by comparing the MCTS iteration rates of a 4-bit adder (a small module), a 64-bit adder (a large module) without modularity,  and a 64-bit adder with modularity. Modularity improves the MCTS iteration rate for the 64-bit adder by $3\times$.
Thus, in summary, we leverage the modular nature of Verilog codes to limit the number of tokens generated for larger modules by reusing the optimized codes generated for the smaller modules by our MCTS formulation.

\subsection{Putting It All Together}\label{sec:putting_it_all_together}
Figure~\ref{fig:final_framework_figure} illustrates the final MCTS framework. We perform the Verilog code generation starting from smaller modules and moving up to the larger modules to reuse the smaller optimized codes (Sec.~\ref{sec:challenge_sol_2}).
We start with hand-designed prompts for our code generation problems.
Depending on whether the prompt is for a large module or not, we incorporate solution 2 to reuse optimized modules found by our MCTS approach for smaller modules.
The prompt is given to our agent, which uses the MCTS algorithm along with LLM to complete code generation. In this process, we utilize our solution 1 to reduce the search space by filtering out tokens corresponding to comment lines/phrases. Once the code generation is done, a reward is computed for the agent (Eq. (\ref{eq:1})), which is backpropagated to all nodes in the current path and used to update the parameters for the MCTS algorithm used in Eq. (\ref{eq:search-policy}). Then, the next episode/iteration begins, and the cycle continues. Eventually, the agent learns to select the optimal path(s), i.e., path(s) that yield the highest reward. Next, we demonstrate the efficacy of MCTS in generating Verilog codes with correct functionality and desired performance.
\begin{figure*}[t]
    \centering
    \includegraphics[width=\textwidth]{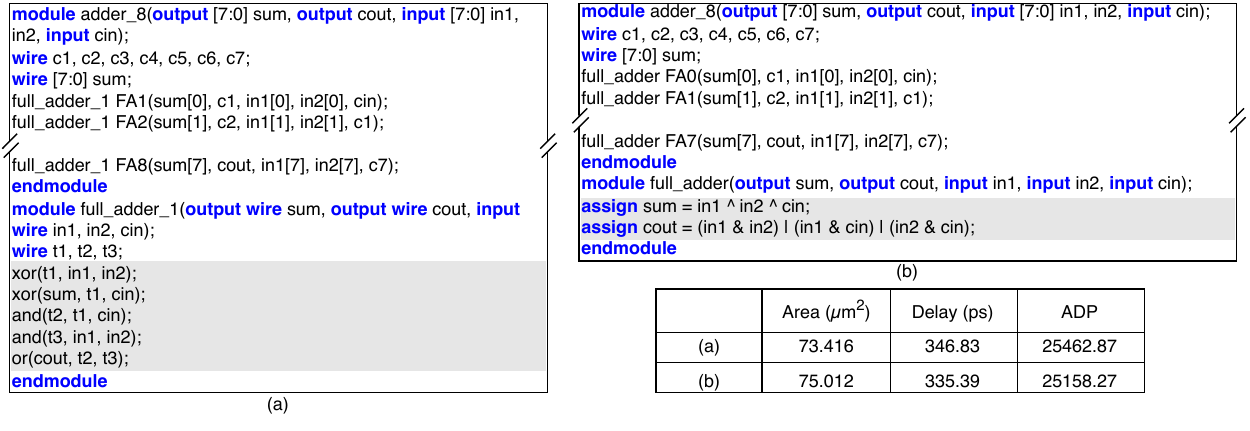}
    \caption{VeriGen+MCTS (a) initial and (b) final optimized functional codes for the 8-bit adder. Lines shaded gray highlight the differences. Note that baseline VeriGen does not produce a functionally correct code, so we do not have such a listing for it.}
    \label{fig:listings}
\end{figure*}
\section{Experimental Evaluation}
\label{sec:results}

\subsection{Experimental Setup}
We use the VeriGen-2B LLM~\cite{thakur_benchmarking_2023} as our baseline LLM as it outperforms chatGPT  (GPT-3.5-turbo) and GPT4 for Verilog code generation.\footnote{Although VeriGen-16B is the best performing VeriGen model~\cite{thakur_benchmarking_2023}, we used VeriGen-2B for our evaluation due to its faster runtime and to demonstrate the potential of MCTS in enhancing LLMs.} VeriGen is obtained by fine-tuning the CodeGen LLM~\cite{nijkamp_conversational_2022} using a Verilog training dataset. We implemented our MCTS algorithm using Python 3.8 and performed all experiments on an NVIDIA RTX A5000 GPU machine with 24 GB RAM. We used \textit{Icarus Verilog 10.3} for checking the compilability of the generated RTL codes. We used custom \textit{Yosys} scripts to synthesize the RTLs into gate-level netlists and to obtain the PPA values, which are used to compute the reward.
We set the values of $\alpha_{NC}$, $\alpha_{NF}$, and $\alpha_B$ as $-1$, $-0.1$, and $0.5$, respectively, in the reward function (Eq. (\ref{eq:1})). 
We created a dataset of 15 Verilog problems consisting of adders, multipliers, and MACs with bit widths in $\{4,8,16,32,64\}$. We chose these problems since they are widely used circuits integral to most modern processors and systems-on-chip. We consider a module to be a large module and use our solution if its bit width is $\geq 32$. 

\subsection{LLM Efficacy Results}
In this section, we first compare the performance of different approaches for LLM-based Verilog code generation in terms of producing functionally correct results. We compare our MCTS-based approach (VeriGen+MCTS) with the baseline VeriGen LLM, which uses greedy search to select the next tokens. We also incorporate Beam Search into VeriGen since it chooses a set of next tokens and explores them all instead of greedily choosing the most likely token. Table~\ref{tab:main_results} demonstrates the results: our VeriGen+MCTS approach produces functionally correct codes for all tested modules, whereas VeriGen and VeriGen+Beam Search fail to generate functionally correct codes for $14$ and $11$ of the $15$ tested modules, respectively. This highlights a critical limitation of existing LLM-based Verilog code generation approaches.

\begin{table}[t]
\centering
\caption{Comparison of performances of Vanilla VeriGen (which uses greedy search), VeriGen+Beam Search, and VeriGen+MCTS (this work) in terms of producing functionally correct code. \textcolor{cadmiumgreen}{\cmark} (\textcolor{red}{\xmark}) indicates successful (failed) generation of functionally correct code. }
\label{tab:main_results}
\resizebox{0.48\textwidth}{!}{
\begin{tabular}{|cc|ccc|}
\hline
 \multicolumn{1}{|c|}{Module} & 
 \multicolumn{1}{c|}{\begin{tabular}[c]{@{}c@{}}Bit\\Width\end{tabular}} & \multicolumn{1}{c|}{\begin{tabular}[c]{@{}c@{}}Vanilla VeriGen\\ (Greedy Search)\end{tabular}} & \multicolumn{1}{c|}{\begin{tabular}[c]{@{}c@{}}VeriGen+\\Beam Search\end{tabular}} & \multicolumn{1}{c|}{\begin{tabular}[c]{@{}c@{}}VeriGen+MCTS\\(this work)\end{tabular}} \\ \hline
\multicolumn{1}{|c|}{\multirow{5}{*}{Adders}} & 4 & \multicolumn{1}{c|}{\textcolor{cadmiumgreen}{\cmark}} & \multicolumn{1}{c|}{\textcolor{cadmiumgreen}{\cmark}} & \textcolor{cadmiumgreen}{\cmark} \\ \cline{2-5} 
\multicolumn{1}{|c|}{} & 8 & \multicolumn{1}{c|}{\textcolor{red}{\xmark}} & \multicolumn{1}{c|}{\textcolor{red}{\xmark}} & \textcolor{cadmiumgreen}{\cmark} \\ \cline{2-5} 
\multicolumn{1}{|c|}{} & 16 & \multicolumn{1}{c|}{\textcolor{red}{\xmark}} & \multicolumn{1}{c|}{\textcolor{cadmiumgreen}{\cmark}} & \textcolor{cadmiumgreen}{\cmark}  \\ \cline{2-5} 
\multicolumn{1}{|c|}{} & 32 & \multicolumn{1}{c|}{\textcolor{red}{\xmark}} & \multicolumn{1}{c|}{\textcolor{cadmiumgreen}{\cmark}} & \multicolumn{1}{c|}{\textcolor{cadmiumgreen}{\cmark}}  \\ \cline{2-5} 
\multicolumn{1}{|c|}{} & 64 & \multicolumn{1}{c|}{\textcolor{red}{\xmark}} & \multicolumn{1}{c|}{\textcolor{red}{\xmark}} &  \multicolumn{1}{c|}{\textcolor{cadmiumgreen}{\cmark}}  \\ \hline
\multicolumn{1}{|c|}{\multirow{5}{*}{Multipliers}} & 4 & \multicolumn{1}{c|}{\textcolor{red}{\xmark}} & \multicolumn{1}{c|}{\textcolor{red}{\xmark}} & \textcolor{cadmiumgreen}{\cmark} \\ \cline{2-5} 
\multicolumn{1}{|c|}{} & 8 & \multicolumn{1}{c|}{\textcolor{red}{\xmark}} & \multicolumn{1}{c|}{\textcolor{red}{\xmark}} & \textcolor{cadmiumgreen}{\cmark} \\ \cline{2-5} 
\multicolumn{1}{|c|}{} & 16 & \multicolumn{1}{c|}{\textcolor{red}{\xmark}} & \multicolumn{1}{c|}{\textcolor{cadmiumgreen}{\cmark}} & \textcolor{cadmiumgreen}{\cmark} \\ \cline{2-5} 
\multicolumn{1}{|c|}{} & 32 & \multicolumn{1}{c|}{\textcolor{red}{\xmark}} & \multicolumn{1}{c|}{\textcolor{red}{\xmark}} & \multicolumn{1}{c|}{\textcolor{cadmiumgreen}{\cmark}}  \\ \cline{2-5} 
\multicolumn{1}{|c|}{} & 64 & \multicolumn{1}{c|}{\textcolor{red}{\xmark}} & \multicolumn{1}{c|}{\textcolor{red}{\xmark}} & \multicolumn{1}{c|}{\textcolor{cadmiumgreen}{\cmark}} \\ \hline
\multicolumn{1}{|c|}{\multirow{5}{*}{MAC Units}} & 4 & \multicolumn{1}{c|}{\textcolor{red}{\xmark}} & \multicolumn{1}{c|}{\textcolor{red}{\xmark}} & \multicolumn{1}{c|}{\textcolor{cadmiumgreen}{\cmark}}  \\ \cline{2-5} 
\multicolumn{1}{|c|}{} & 8 & \multicolumn{1}{c|}{\textcolor{red}{\xmark}} & \multicolumn{1}{c|}{\textcolor{red}{\xmark}} & \multicolumn{1}{c|}{\textcolor{cadmiumgreen}{\cmark}}  \\ \cline{2-5} 
\multicolumn{1}{|c|}{} & 16 & \multicolumn{1}{c|}{\textcolor{red}{\xmark}} & \multicolumn{1}{c|}{\textcolor{red}{\xmark}} & \multicolumn{1}{c|}{\textcolor{cadmiumgreen}{\cmark}}  \\ \cline{2-5} 
\multicolumn{1}{|c|}{} & 32 & \multicolumn{1}{c|}{\textcolor{red}{\xmark}} & \multicolumn{1}{c|}{\textcolor{red}{\xmark}} & \multicolumn{1}{c|}{\textcolor{cadmiumgreen}{\cmark}}  \\ \cline{2-5} 
\multicolumn{1}{|c|}{} & 64 & \multicolumn{1}{c|}{\textcolor{red}{\xmark}} & \multicolumn{1}{c|}{\textcolor{red}{\xmark}} & \multicolumn{1}{c|}{\textcolor{cadmiumgreen}{\cmark}} \\ \hline
\end{tabular}
}
\end{table}

\begin{table*}[t]
\centering
\caption{Comparison of area-delay product (ADP) results for Vanilla VeriGen, VeriGen+Beam Search (BS), and VeriGen+MCTS. Since VeriGen and VeriGen+BS do not result in functionally correct codes for most of the modules, we denote these as ``N/A''.}
\label{tab:main_results_2}
\resizebox{\textwidth}{!}{
\begin{tabular}{|c|c|ccccc|ccccc|ccccc|}
\hline
\multicolumn{2}{|c|}{Module} & \multicolumn{5}{c|}{Adders} & \multicolumn{5}{c|}{Multipliers} & \multicolumn{5}{c|}{MAC Units} \\ \hline
\multicolumn{2}{|c|}{Bit Width} & \multicolumn{1}{c|}{4} & \multicolumn{1}{c|}{8} & \multicolumn{1}{c|}{16} & \multicolumn{1}{c|}{32} & 64 & \multicolumn{1}{c|}{4} & \multicolumn{1}{c|}{8} & \multicolumn{1}{c|}{16} & \multicolumn{1}{c|}{32} & 64 & \multicolumn{1}{c|}{4} & \multicolumn{1}{c|}{8} & \multicolumn{1}{c|}{16} & \multicolumn{1}{c|}{32} & 64 \\ \hline
Vanilla VeriGen & ADP ($\times 10^3$)& \multicolumn{1}{c|}{8.08} & \multicolumn{1}{c|}{N/A} & \multicolumn{1}{c|}{N/A} & \multicolumn{1}{c|}{N/A} & N/A & \multicolumn{1}{c|}{N/A} & \multicolumn{1}{c|}{N/A} & \multicolumn{1}{c|}{N/A} & \multicolumn{1}{c|}{N/A} & N/A & \multicolumn{1}{c|}{N/A} & \multicolumn{1}{c|}{N/A} & \multicolumn{1}{c|}{N/A} & \multicolumn{1}{c|}{N/A} & N/A \\ \hline
VeriGen+BS & ADP ($\times 10^3$)& \multicolumn{1}{c|}{8.08} & \multicolumn{1}{c|}{N/A} & \multicolumn{1}{c|}{138.47} & \multicolumn{1}{c|}{459.32} & N/A & \multicolumn{1}{c|}{N/A} & \multicolumn{1}{c|}{N/A} & \multicolumn{1}{c|}{2843.16} & \multicolumn{1}{c|}{N/A} & N/A & \multicolumn{1}{c|}{N/A} & \multicolumn{1}{c|}{N/A} & \multicolumn{1}{c|}{N/A} & \multicolumn{1}{c|}{N/A} & N/A \\ \hline
\multirow{3}{*}{\begin{tabular}[c]{@{}c@{}}VeriGen+MCTS\\ (this work)\end{tabular}} & \multicolumn{1}{c|}{ADP ($\times 10^3$)} & \multicolumn{1}{c|}{7.62} & \multicolumn{1}{c|}{25.16} & \multicolumn{1}{c|}{94.39} & \multicolumn{1}{c|}{368.86} & 1441.20 & \multicolumn{1}{c|}{41.27} & \multicolumn{1}{c|}{506.74} & \multicolumn{1}{c|}{2843.16} & \multicolumn{1}{c|}{20173.17} & 144831.83 & \multicolumn{1}{c|}{128.51} & \multicolumn{1}{c|}{668.04} & \multicolumn{1}{c|}{3623.44} & \multicolumn{1}{c|}{23516.33} & 174361.19\\ \cline{2-17}
& \multicolumn{1}{c|}{Impr./VeriGen} & \multicolumn{1}{c|}{5.69\%} & \multicolumn{1}{c|}{N/A} & \multicolumn{1}{c|}{N/A} & \multicolumn{1}{c|}{N/A} & N/A & \multicolumn{1}{c|}{N/A} & \multicolumn{1}{c|}{N/A} & \multicolumn{1}{c|}{N/A} & \multicolumn{1}{c|}{N/A} & N/A & \multicolumn{1}{c|}{N/A} & \multicolumn{1}{c|}{N/A} & \multicolumn{1}{c|}{N/A} & \multicolumn{1}{c|}{N/A} & N/A \\ \cline{2-17}
& \multicolumn{1}{c|}{Impr./VeriGen+BS} & \multicolumn{1}{c|}{5.69\%} & \multicolumn{1}{c|}{N/A} & \multicolumn{1}{c|}{31.8\%} & \multicolumn{1}{c|}{19.6\%} & N/A & \multicolumn{1}{c|}{N/A} & \multicolumn{1}{c|}{N/A} & \multicolumn{1}{c|}{0.00\%} & \multicolumn{1}{c|}{N/A} & N/A & \multicolumn{1}{c|}{N/A} & \multicolumn{1}{c|}{N/A} & \multicolumn{1}{c|}{N/A} & \multicolumn{1}{c|}{N/A} & N/A \\ \hline
\end{tabular}
}
\end{table*}

Next, we analyze the PPA improvement obtained by VeriGen+ MCTS. Table~\ref{tab:main_results_2} compares the area-delay product (ADP) of our approach with the existing approaches. Since these are unable to produce a functionally correct code for most of the modules, we denote those as ``N/A''. For the other modules, it is evident that our approach yields codes with a significant improvement in ADP (average improvement of 5.69\% and 14.27\% over VeriGen and VeriGen+Beam Search, respectively). Additionally, we also show an example of how our VeriGen+MCTS approach optimizes the 8-bit adder to obtain a minimal ADP in Figure~\ref{fig:listings}.
The highlighted lines in the codes show the differences that result in different ADPs.

\begin{figure}[t]
    \centering
    \includegraphics[width=0.48\textwidth]{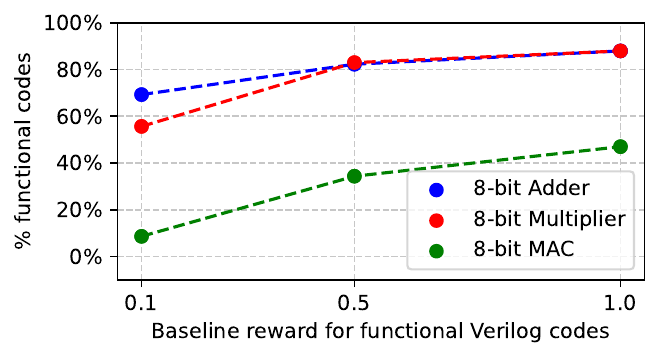}
    \caption{Impact of the baseline reward}
    \label{fig:impact_of_baseline_reward}
\end{figure}

Now, we analyze the impact of changing the baseline reward value, $\alpha_B$ in Eq. (\ref{eq:1}), on the performance of our VeriGen+MCTS approach (Figure~\ref{fig:impact_of_baseline_reward}).
As the baseline reward increases, the percentage of functional codes increases. This is because having a higher baseline reward ($\alpha_B$) results in that term dominating over the $1-\frac{a\times d}{a^{\dagger}\times d^{\dagger}}$ term in the reward equation (Eq. (\ref{eq:1})), resulting in a limited exploration by the agent.
This can be observed from the fact that for baseline reward values of $0.1$, $0.5$, and $1.0$, the longest sequence with the same ADP value for the 8-bit adder is $7$, $10$, and $25$, respectively, meaning that for larger values of baseline reward, the agent tends to exploit more.
So, an intermediate value of $\alpha_B$ is better for a good balance of exploration and exploitation.

\begin{figure}[t]
    \centering
    \includegraphics[width=0.48\textwidth]{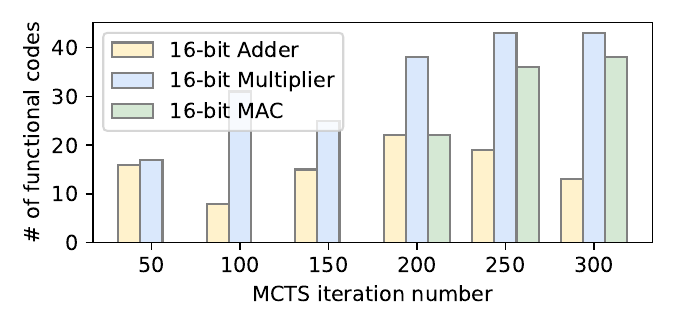}
    \caption{Impact of the number of MCTS iterations}
    \label{fig:impact_of_MCTS_iterations}
\end{figure}

We also analyze the impact of the number of MCTS iterations on the number of functionally correct codes for the 16-bit adder, multiplier, and MAC unit in Figure~\ref{fig:impact_of_MCTS_iterations}.
Our VeriGen+MCTS approach learns to generate functionally correct codes for adder and multiplier within $50$ iterations, whereas the MAC unit requires $\approx200$ iterations. This is because the MAC unit is more complicated and requires more tokens than the adder and multiplier, meaning that the size of the tree to be explored by the agent is larger for the MAC unit. Hence, the agent requires more iterations to learn to generate compilable and then functional results for the MAC unit.

\definecolor{lightGrey}{rgb}{0.9, 0.9, 0.9}
\colorlet{pink}{red!40}
\newcommand{\Hilight}{\makebox[0pt][l]{\color{lightGrey}\rule[-0.5ex]{\linewidth}{2.5ex}}}

\section{Discussion and Future Work}
\label{sec:discussion}
In this work, we propose using MCTS to solve the MDP for Verilog code generation by exploring trees of tokens generated by an LLM. Although this approach results in functional as well as optimized codes, it is time-intensive since the exploration of the tree needs to be done for each new module to be generated. A potential approach to alleviate this problem could be to fine-tune the LLM (i.e., update its parameters) using the rewards to generate desired PPA-optimized code. However, fine-tuning usually requires a large amount of training data and training time. Thus, future work also needs to strike a balance in the trade-off between fine-tuning and our MCTS-based approach in terms of time and resources required for training 
vs. during inference for each new module. 
\section{Conclusion}
\label{sec:Conclusion}
Recent works on LLMs for Verilog code generation have shown great promise, even outperforming chatGPT  (GPT-3.5-turbo and GPT4).
However, these prior works face challenges such as compilation failure and lack of PPA-optimized code generation.
To address these limitations, we have presented a novel technique to generate optimized Verilog RTL codes using LLMs and MCTS. 
To this end, we addressed the challenges of integrating MCTS with LLMs, such as search efficiency and scalability. We have demonstrated the effectiveness of our technique on various ubiquitous designs such as adders, multipliers, and MAC units of different sizes. Experimental results show that our technique can generate Verilog codes that are functionally correct and optimized for a user-defined objective.
We have also compared our technique with prior works that use LLMs alone or with other search methods and found that our technique outperforms prior works in terms of accuracy and quality. For the largest design generated by VeriGen (16-bit adder w/ beam search), our technique is able to achieve a 31.8\% improvement in the area-delay product.
Our technique can be further improved or augmented by fine-tuning to update the LLM parameters.

\section*{Acknowledgement}
The authors acknowledge the support from the Purdue Center for Secure Microelectronics Ecosystem – CSME\#210205. This work was also partially supported by the National Science Foundation (NSF CNS--1822848 and NSF DGE--2039610) and in part by a Synopsys gift. Any opinions, findings, conclusions, or recommendations expressed herein are those of the authors and do not necessarily reflect those of the funding agencies.

\bibliographystyle{ACM-Reference-Format}
\bibliography{main}

\end{document}